\documentclass{isprs} 
\usepackage{setspace}
\usepackage{geometry} 
\usepackage{epstopdf}
\usepackage[labelsep=period]{caption}  
\usepackage[british]{babel} 
\usepackage[hang]{footmisc}
\usepackage[authoryear]{natbib}
\usepackage{placeins}
\usepackage{here}
\usepackage{xcolor}
\usepackage{subfig}
\usepackage{amsmath}
\usepackage{url}
\usepackage[colorlinks=true]{hyperref}
\hypersetup{
citecolor=[rgb]{0.0, 0.53, 0.74}
}

\begin{document}

\title{Learning to Sieve: Prediction of Grading Curves from Images of Concrete Aggregate}
\date{}

\author{
Max Coenen\textsuperscript{1,}\thanks{Corresponding author}\ , Dries Beyer\textsuperscript{1}, Christian Heipke\textsuperscript{2}, Michael Haist\textsuperscript{1}
}

\address{
	\textsuperscript{1 } Institute of Building Materials Science, Leibniz University Hannover, Germany\\
	(m.coenen, d.beyer, haist)@baustoff.uni-hannover.de\\
	\textsuperscript{2 } Institute of Photogrammetry and GeoInformation, Leibniz University Hannover, Germany\\
	heipke@ipi.uni-hannover.de\\
	\ \\ 
	
}

\icwg{}   

\abstract{A large component of the building material concrete consists of aggregate with varying particle sizes between 0.125 and 32 mm. 
Its actual size distribution significantly affects the quality characteristics of the final concrete in both, the fresh and hardened states. 
The usually unknown variations in the size distribution of the aggregate particles, which can be large especially when using recycled aggregate materials, are typically compensated by an increased usage of cement which, however, has severe negative impacts on economical and ecological aspects of the concrete production. 
In order to allow a precise control of the target properties of the concrete, unknown variations in the size distribution have to be quantified to enable a proper adaptation of the concrete's mixture design in real time. To this end, this paper proposes a deep learning based method for the determination of concrete aggregate grading curves. 
In this context, we propose a network architecture applying multi-scale feature extraction modules in order to handle the strongly diverse object sizes of the particles.  
Furthermore, we propose and publish a novel dataset of concrete aggregate used for the quantitative evaluation of our method.
}

\keywords{Granulometry, concrete aggregate, particle size distribution, automation in construction, deep learning}

\maketitle

\sloppy

\section{Introduction}\label{sec:introduction}
\begin{figure}[ht]
\centering
\includegraphics[width=1.0\columnwidth]{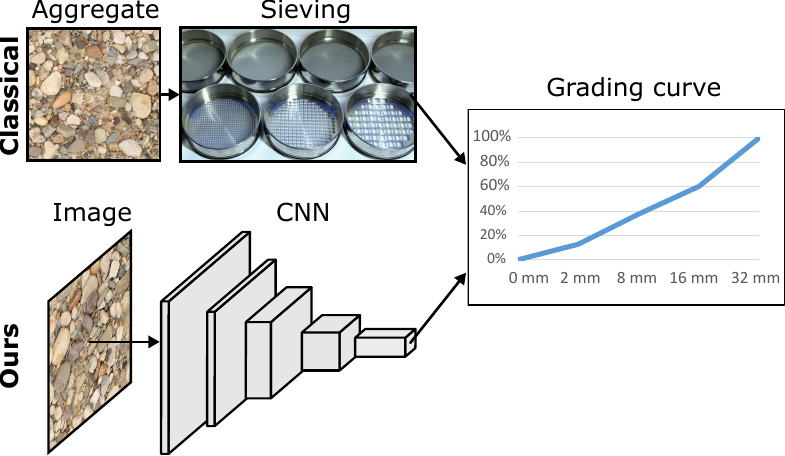}
\caption{Traditional estimation of the grading curve using mechanical sieving (top row) vs.\;image and deep learning based prediction of grading curves as proposed here (bottom row).}
\label{fig:intro} 
\end{figure}

Concrete is one if the most used building materials worldwide. 
With up to 80\% of volume, a large constituent of concrete consists of fine and coarse aggregate particles (normally, sizes of 0.1\,mm to 32\,mm) which are dispersed in a cement paste matrix. The size distribution of the aggregates (i.e.\;the grading curve) substantially affects the properties and quality characteristics of concrete, such as e.g.\;its workability at the fresh state and the mechanical properties at the hardened state.
In practice, usually the size distribution of small samples of the aggregate is determined by manual mechanical sieving (cf.\;top of Fig.~\ref{fig:intro}) and is considered as representative for a large amount of aggregate.
However, the size distribution of the actual aggregate used for individual production batches of concrete varies, especially when e.g.\;recycled material is used as aggregate.
As a consequence, the unknown variations of the particle size distribution have a negative effect on the robustness and the quality of the final concrete produced from the raw material. 
These variations are usually compensated by a distinct increase of the cement proportion, which, however, is neither economically nor ecologically justifiable.
Towards the goal of deriving precise knowledge about the actual particle size distribution of the aggregate, thus eliminating the unknown variations in the material's properties, we propose an approach for the image based prediction of the size distribution of concrete aggregates (cf.\;bottom of Fig.~\ref{fig:intro}).
Incorporating such an approach into the production chain of concrete enables to react on detected variations in the size distribution of the aggregate in real-time by adapting the composition, i.e.\;the mixture design of the concrete accordingly, so that the desired concrete properties are reached \citep{Haist2022}.
%
%

Towards this goal, we make the following contributions:\\
\textbf{1)} We present a deep learning architecture in form of a convolutional neural network (CNN), denoted as \textit{Aggregate Network} (AggNet), for the prediction of the grading curve from images of concrete aggregate. Particularly, we propose a multi-scale feature encoder as substantial part of the network, that is able to handle the large range of different particle sizes (fractions of 0-2\;mm up to 32\;mm), leading to an improved performance. \\
\textbf{2)} We present a new data set of high-resolution images of raw aggregate material, each image associated with the ground truth grading curve obtained from reference measurements, i.e.\;from mechanical sieving. 
The images are acquired and made publicly available during the course of this paper with a ground sampling distance (GSD) of 8 [px/mm] (although in the experiments of this paper, a lower GSD is used). \\
\textbf{3)} In an extensive evaluation, we investigate the influence of the image resolution, i.e.\;the GSD, and of the individual constituents of the proposed method on the prediction quality. Furthermore, we analyse the patterns in the correct and incorrect predictions made by the network to derive deeper insights into difficulties and limitations w.r.t.\;the ability to differentiate different particle size distributions. In this context, we also compare the behaviour of the network to the classification ability of human experts in order to interpret and to explain qualities and limitations in the performance.

The paper is organised as follows.
An overview on related work is given in Sec.~\ref{sec:relatedWork}. 
The proposed method for the prediction of grading curves is explained in Sec.~\ref{sec:methodology}.
In Sec.~\ref{sec:testsetup}, our data set and the evaluation strategy is presented. The evaluation of the results is done in Sec.~\ref{sec:evaluation} and the paper closes with a conclusion in Sec.~\ref{sec:conclusion}.

\section{Related work} \label{sec:relatedWork}
Aggregate  has  a  considerable  influence  on  the  properties of fresh concrete (e.g.\;workability,  consistency,  mix stability) and hardened concrete (e.g.\;strength, durability).
In  addition  to  technical  parameters  such  as e.g.\;density or particle shape, particle size in particular is of decisive importance for concrete production.
In this context, a specific adjustment of the aggregate size distribution - also known as grading curve - is critical in order to fill the available space as densely as possible with aggregate. 
In practice, the grading curve of the aggregate is determined from random samples using a sieving method for the determination of the particle size proportions.
In this process, a very small quantity of aggregate - a few kilograms - is representative for many tons of aggregate used for a variety of concrete mixes, therefore, not taking into account the very wide range of material scatter that is typically inherited by aggregates.
In this work, an image based classification of aggregate grading curves is proposed, which delivers a real-time capable analysis of the size distribution of concrete aggregate.
Establishing this approach as an online measurement process by installing cameras over the aggregate feeding belt, thus, observing the total amount of aggregates actually used for the particular concrete mix, enables the real-time opportunity of adapting the concrete mix design accordingly (we refer the reader to \citep{Haist2022}, where a concept for the online concrete production control is proposed).

The estimation of object size distributions from images has a wide field of interest, ranging from applications in geosciences \citep{Buscombe2020}, biological and medical applications \citep{Sharma2020}, over hydrological \citep{Lang2021} and geographical \citep{Soloy2020} applications, to industrial practices \citep{ Hamzeloo2014}.
One line of work towards determining size distributions follows an \textbf{object-based procedure}, in which individual objects are segmented first, and their size distribution is computed from the segmentations in a second step.
\citet{Kumara2012} used grayscale thresholding and morphological operators for the segmentation of gravel. 
In \citep{Hamzeloo2014}, the segmentation of aggregate is based on edge detections. \citet{Lira2006} and \citet{Terzi2017} applied watershed transformation for particle segmentation. 
However, in all approaches mentioned so far, often extensive manual parameter tuning is necessary, which restricts the application to large and diverse sets of images and leads to difficulties for scenes of low contrast and complex textures.
\citet{Soloy2020} avoid the requirements of manual intervention in the segmentation process by learning a convolutional neural network (CNN) using the MaskRCNN \citep{maskRCNN} architecture for the instance segmentation of grains on pebble beaches.
However, object based approaches in general suffer from several difficulties. 
On the one hand, they are sensitive to partial occlusions and require a large enough image resolution to enable the segmentation of individual particles. 
As a consequence, the estimation of size distributions of small particle size fractions from scenes with a coarse image resolution, i.e.\;of small grains that are not visible by the naked eye, is not possible.
Besides, in most applications, the volumetric size distribution of the objects is the target of interest. 
Since an instance segmentation delivers objects sizes only in 2D, i.e.\;the size of the two-dimensional projection of the particles, an explicit conversion to the volumetric distribution is necessary. 
\citet{Buscombe2013} proposes a statistical model for the area-to-volume/mass conversion.
In \citep{Hamzeloo2014}, the conversion is done by assuming an ellipsoidal or spherical shape model for the particles. 

In contrast to the described object-based procedure, \textbf{statistical approaches} avoid the explicit detection and modelling of individual objects by relying on global image statistics. 
In order to overcome the limitations of object-centred approaches, statistical methods aim at predicting the size distribution directly from the raw image. 
In this context, \citet{Olivier2019} propose to learn a CNN for the image based size characterisation of ore.
However, they only distinguish between classes of ore samples containing or not containing large rocks, and do not deliver size distributions.
In \citep{Buscombe2020}, a CNN is proposed that predicts the percentiles of the cumulative grading curve of sediments. 
However, predicting the cumulative representation of the grading curve using a linear activation as final layer leads to an ill-constrained definition of the output.
In contrast, \citet{Olivier2020} predict the size distribution as proportional histogram and introduce a loss function which tries to enforce the output to sum up to one. 
Similarly, \citet{Lang2021} treat the size distribution as a probability distribution and a Softmax function is used as final activation to ensure a valid output. 
However, their generation of training data requires the manual annotation of instance masks of individual grains and from these annotations, a quasi-sieve throughput is computed using a statistical model. 
In \citep{Sharma2020}, dot annotations of individual objects are required for the generation of training data, used to train a CNN to predict the size distribution of fly larvae. 
As a consequence, the latter two authors rely on image resolution which allows the identification if individual object instances.
In our paper, we avoid both, the requirement of image-level annotations and of volumetric modelling, as well as of the requirement of an image resolution that is large enough to identify each individual particle. 
Instead, we design aggregate samples with known mass distributions and train a network to predict the mass grading curves, while therefore, implicitly learning the conversion from the 2D projection space to the volumetric/mass proportions. 
Furthermore, our method learns the prediction of grading curves with a wide spectrum of particle sizes (0-32\;mm), including very fine particles whose individual identification in the images is impossible. 

\section{Methodology} \label{sec:methodology}

In this work, we propose an approach to automatically determine the grading curve of aggregates used for the production of concrete.
A formal definition of the problem and the goal of this paper is given in Sec.~\ref{sec:problemStatement}.
Requirements on the data and necessary preprocessing steps are explained in Sec.~\ref{sec:preprocessing}.
Finally, a technical description of our proposed method for the deep learning based estimation of the particle size distribution from images is provided in Sec.~\ref{sec:CNNMethod}.

\subsection{Formal problem definition} \label{sec:problemStatement}
Given an image $\mathbf{I}$ depicting raw material of concrete aggregates, the goal of this work is to automatically determine the grain size distribution, also denoted as grading curve, of the shown material. 
Per definition, the grading curve is a histogram (often also depicted as a cumulative histogram) in which grain size intervals (bins) are represented on the abscissa (x-axis) and the quantity proportion (or cumulative quantity) is shown on the ordinate (y-axis).
Note that both forms, the histogram and the cumulative histogram, are redundant representations as they can directly be computed from each other.

More formally, in this paper, we aim at deriving a mapping function $f: \mathbf{I} \rightarrow \mathcal{G}$ which maps the image $\mathbf{I}$ to the grading curve $\mathcal{G}$. 
In this context, we propose to represent $f$ by a convolutional neural network (CNN) which takes the image as input and is trained to predict the corresponding grading curve as output. 
To this end, we introduce a discrete representation of the grading curve $\mathcal{G}$ in this paper.
In this representation, the grading curve $\mathcal{G} = C_i$ is parametrised by a discrete variable $C_i \in \mathbf{C}$ corresponding to one of a set of $i =1...N$ predefined grading curves $\mathbf{C}$. 
One possible way of declaring the set of grading curves is e.g.\;to define them according to the limiting grading curves of DIN 1045.
As a consequence of this representation, determining the discrete form of the grading curve from the image corresponds to a classification problem, in which the variables in $\mathbf{C}$ form the set of classes and the image is to be classified as one of them.

\subsection{Image preprocessing} \label{sec:preprocessing}
As just described, the goal of this work is the prediction of the size distribution of 3D objects, in particular of aggregate particles, from their perspective projection into an image. 
In this context, it is important to consider that distances/sizes in object space are mapped to distances/sizes in image space, whose scale factor can vary greatly from image point to image point, depending on the structure of the 3D scene and/or the viewpoint of the image.
Only for a planar 3D scene and a strict nadir view of the camera, the scale factor (also denoted as ground sampling distance GSD) is constant over the whole image.
We argue, that for the task of estimating the particle size distribution of objects, a constant GSD is crucial, since unknown and non-constant scales lead to ambiguities for that task.
In order to achieve the property of a constant GSD, we assume that the surface of the aggregate material is planar and, therefore, can approximately be represented by a plane. 
Following this assumption, the projective transformation between the planar aggregate surface in 3D object space and the 2D image plane can mathematically be described by a homography $\mathcal{H}$.
Given this homography, the image can be transformed in a way that it corresponds to a nadir projection of the aggregate surface, i.e.\,it is perspectively rectified.
The computation of $\mathcal{H}$ requires at least four homologous points, i.e.\;points whose coordinates are known and/or observed on both planes.
We assume that enough homologous point are available in the setup (e.g.\;by installing reference markers with known coordinates in a local coordinate system which lies in the same plane as the aggregate surface), enabling the computation of $\mathcal{H}$ and consequently, enabling the perspective rectification of the image $\mathbf{I}$ with a known and constant GSD.
In this way, we do not restrict the acquisition geometry to a strict nadir setup but, instead, allow the usage of arbitrary viewpoints of the camera providing a higher flexibility w.r.t.\; the acquisition requirement.

\subsection{Deep learning based prediction of particle grading curves} \label{sec:CNNMethod}
In this work, we aim at finding a functional relationship
\begin{equation}
\mathcal{G} = f(\mathbf{I}, \mathbf{w}) \label{eq:cnnfunction}
\end{equation}
which determines the grading curve $\mathcal{G}$ from a given image $\mathbf{I}$. 
In this context, the function $f(\cdot)$ is represented by a CNN and, thus, is defined by the architecture of that network. 
In order to derive the function parameters $\mathbf{w}$, which are the network parameters, namely the weight and biases of the the network's filter kernels, training data is used in order to learn the parameters by minimising a dedicated loss function. 
Details on the architecture (Sec.~\ref{sec:architecture}) and on the training procedure (Sec.~\ref{sec:training}) of the CNN are described in the following sections.

\subsubsection{CNN Architecture} \label{sec:architecture}
\begin{figure}[ht]
\centering
\includegraphics[width=0.9\columnwidth]{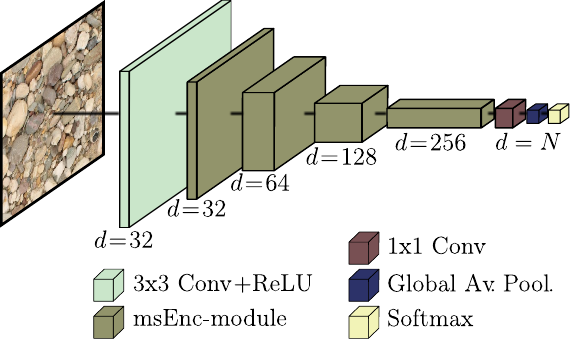}
\caption{High-level overview on the \textit{AggNet} architecture proposed in this work for grading curve prediction.}
\label{fig:architecture} 
\end{figure}

A high-level overview on the particular architecture of the CNN that we propose for the task of grading curve estimation, denoted as \textit{Aggregate Network} (AggNet) in the remainder of this paper, is shown in Fig.~\ref{fig:architecture}.
%
%

The input to our CNN is a three-channel colour image showing a pile of aggregate particles, perspectively rectified as described in Sec.~\ref{sec:preprocessing}.
We propose a \textit{multi-scale residual feature encoder} module (msEnc-module) in this paper and, after an initial convolutional layer with $d=32$ filter kernels of size 3x3, we  apply a series of four such modules to the network pipeline. \\

\begin{figure}[ht]
\centering
\includegraphics[width=0.9\columnwidth]{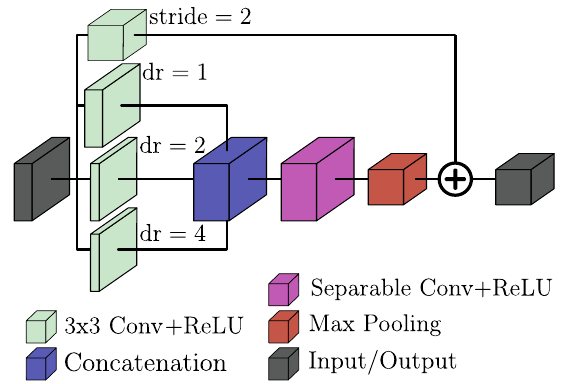}
\caption{Overview on the architecure of the proposed \textit{msEnc}-module. The applied dilation rate $dr$ of the multi-scale layers is shown in the figure.}
\label{fig:msEnc} 
\end{figure}

\textbf{msEnc-module:} As can be seen in Fig.~\ref{fig:architecture}, each of the msEnc-modules takes a feature map of size $m\times n$ as input and produces a a feature map of depth $d$ and of a spatial size $m/2 \times n/2$, i.e. which is half the size of the respective module's input.
A detailed overview on the msEnc-module's intrinsic architecture is shown in Fig.~\ref{fig:msEnc}.
Inside each module, two intermediate representations are computed from the initial feature map. 
The first representation is produced by a convolutional layer using a kernel size of 3x3 and a stride of 2 and serves as residual feature map \citep{ResNet2016}.
The second representation is computed by a concatenation of three multi-scale convolutional layers followed by a depthwise separable convolutional layer \citep{xception2017,MobileNet} and downsampling using max pooling with kernel size 2x2 and stride 2. 
Compared to the standard convolutional layers, depthwise separable convolutional layers have the effect of drastically reducing computational cost and model size while preserving the ability of extracting discriminative feature maps and therefore, preserving the quality of performance.
For the multi-scale feature extraction, we apply three dilated convolution layers \citep{DilatedConv2016} with a kernel size of 3x3 and a dilation rate of 1, 2, and 4 respectively.
We argue that the latter multi-scale convolutional filters are especially beneficial for the tackled task of determining the size distribution of aggregate particles whose size can vary drastically.
For instance, in concrete production, usually aggregate fractions from 0-2\;mm up to 32\;mm are incorporated.
As will be shown in the experimental part of this paper (cf. Sec.~\ref{sec:testsetup}), we aim at determining the mass proportions of particles with a size of 1\;mm at the smallest scale and with a size of 32\;mm at the largest scale, which results in a difference in size of factor 32 for the range of objects to be considered here. 
In order to consider these differences in object scale, we believe that the usage of dilated convolutions with different dilation rates allow the extraction of multi-scale features maps that are able to deliver latent representations suitable for both, smaller and larger sized particles at the same time.\\

%
%

After the series of msEnc modules, a convolutional layer using a kernel size of 1 is applied (cf. Fig.~\ref{fig:architecture}).
The number of filters $d$ in this layer corresponds to the number of classes $N$, i.e. the number of predefined grading curves that are to be distinguished.
As a consequence, the produced feature map of that layer has a number of $N$ channels. 
Global average pooling is applied to produce the initial raw output scores $s_{i}$ from that feature map with $i=[1,N]$ as real numbers ranging from $[-\infty, +\infty]$.
Note that in contrast to a fully connected layer, which is usually applied at the end of a classification/regression network and which restricts the applicability of the network to a specific spatial size of the input image, global average pooling renders the network fully convolutional and, therefore, allows to feed images of arbitrary size to the network. 
We point out, that in the application case of this paper, where the network is used to process images acquired on aggregate conveyor belts, this flexibility delivers independence of the method from the size/width of the conveyor belt, as the image size can be adapted accordingly in order to cover the whole width of the belt to be considered for the estimation of the grading curve.
As final operation, a softmax function is used to normalise the raw scores with
\begin{equation}
\hat{s}_{i} = \frac{\exp(s_{i})}{\sum_{i} \exp(s_{i})},
\end{equation}
so that as a consequence, the sum of the final output of the CNN results to $\sum_{i} = 1$. 
In case of the discrete representation $\mathcal{G}$, the output of the CNN can be interpreted as predicted probabilities for each of the classes $C_i \in \mathbf{C}$. 

\subsubsection{Training} \label{sec:training}
Learning all network parameters $\mathbf{w}$ of Eq.~\ref{eq:cnnfunction}, namely all weights and biases of the filter kernels, is done during training by minimising a loss function $\mathcal{L}(\mathbf{w})$. 
Training is performed in a supervised manner, which means that the availability of training samples with corresponding reference data is required.
Starting from a (random) initialisation of the weights, the optimisation is performed iteratively using stochastic mini-batch gradient descent (SGD) \citep{Goodfellow2016}. 
In this work, the \textit{He} initialiser \citep{He2015} is used for weight initialisation and the \textit{Adam} optimiser \citep{Adam2015} is used for weight optimisation.
In case of the proposed discrete representation $\mathcal{G}$ of the grading curve, the CNN corresponds to a multi-class classification network whose training requires training data including a provided reference class for each training sample. 
A frequently used loss for classification tasks, which is also applied here, is the cross-entropy (CE) loss. 
With $N$ different classes to distinguish, the cross-entropy loss $\mathcal{L}_{CE}$ is calculated according to
\begin{equation}
\mathcal{L}_{CE}(\mathbf{w}) = - \sum_{i=1}^{N} b_{i} \log(\hat{s}_{i}),
\end{equation}
where $\hat{s}_i$ denotes the softmax output of the sample for the $i^\text{th}$ class and $b_{i} \in \{0,1\}$ is a binary variable indicating whether the sample belongs to the $i^\text{th}$ class $C_i$ or not.

\textbf{Augmentation:} 
In order to gain robustness and to achieve a better generalisation ability of the CNN model, we propose to apply strong data augmentation of both kinds, geometric and radiometric augmentations, during training. 
More specifically, we make use of random horizontal and vertical flips of the image as geometric augmentations. 
Furthermore, we apply linear and non-linear augmentations to the gray-values of the image with
\begin{equation}
\mathbf{I}^\text{aug} = (\mathbf{I} \cdot \alpha + \beta)^\gamma.
\end{equation}
In this equation, $\alpha$ mainly affects the contrast of the image, where $\alpha > 1$ results in stronger contrast and $\alpha < 1$ causes less contrast. 
The parameter $\beta$ changes the brightness of the image and defining a value for $\gamma \neq 1$ corresponds to a gamma correction applied to the image. 
Furthermore, we apply random colour alterations to the images by applying hue shift, i.e. by shifting the hue channel of the images' HSI colour space using a random offset $\tau$. 

\section{Experimental setup} \label{sec:testsetup}

\subsection{Test data}
%
\begin{figure*}[ht]
\centering
\subfloat[Histogram of the mass-percentage of each grain size fractions and for each class.] {\includegraphics[width=0.45\textwidth]{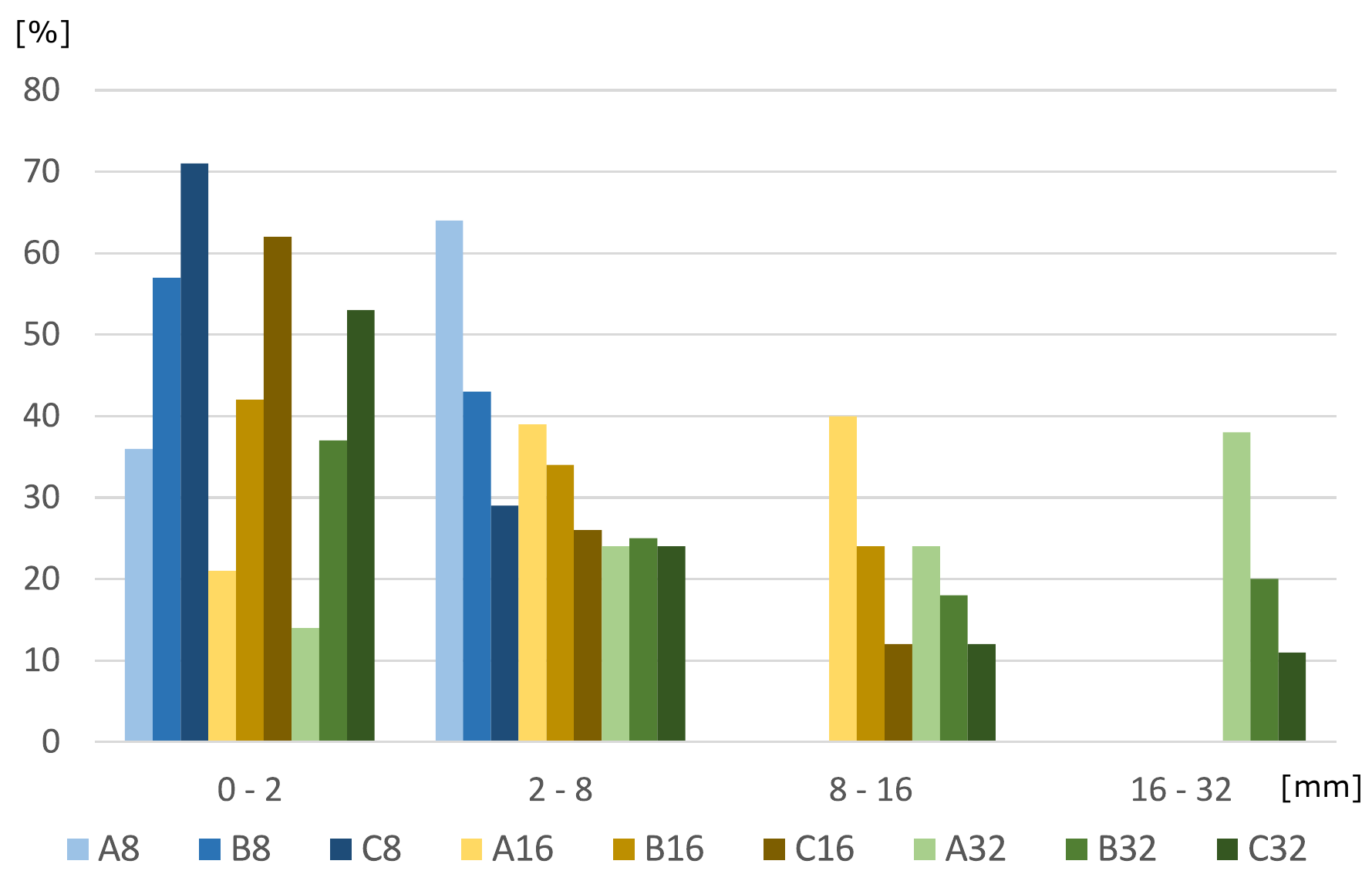}\label{fig:hist01}} 
\hspace{0.5cm}
\subfloat[Cumulative histogram of the grain size distribution for each class.] {\includegraphics[width=0.45\textwidth]{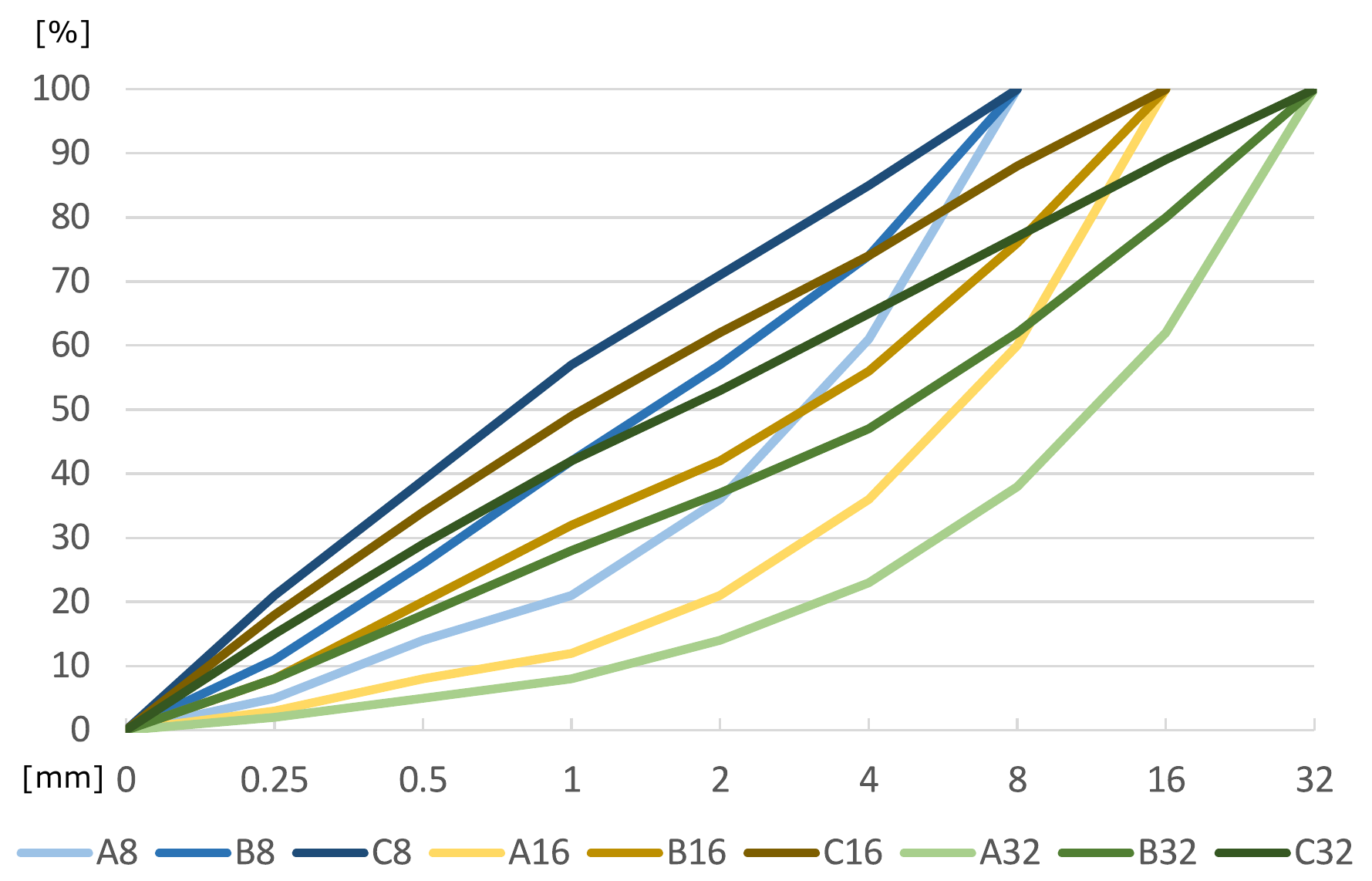}\label{fig:hist02}} 
\caption{Quantitative definition of the nine grading curve classes which are distinguished in this work (vertical axes represent the mass-percentages of the grain size distribution).}
\label{fig:histograms} 
\end{figure*}
%
In order to evaluate the proposed approach for the classification of concrete aggregate grading curves we created a new data set which has been made publicly available in the course of this paper\footnote{The Visual Granulometry Benchmark:\\ \url{https://doi.org/10.25835/etbkk0pb}.}.
More specifically, in our data set, we distinguish between nine different grading curves, each exhibiting a different grain size distribution. 
To this end, we  follow the definition of the limiting grading curves considered in the normative regulations of DIN 1045. 
As a consequence, the number of classes contained in the label set $\mathbf{C}$ and being distinguished in the data set and the experiments of this paper corresponds to $N=9$.
The nine grading curves differ in their maximum particle size (8, 16, or 32\;mm) and in the distribution of the particle size fractions allowing a categorisation of the curves to coarse-grained (A), medium-grained (B) and fine-grained (C) curves, respectively.
A quantitative description of the grain size distribution of the nine curves distinguished in this paper is shown in Fig.~\ref{fig:histograms}, where Fig.~\ref{fig:hist01} contains a histogram of the particle size fractions 0-2, 2-8, 8-16, and 16-32\;mm and Fig.~\ref{fig:hist02} shows the cumulative histograms of the grading curves (the vertical axes represent the mass-percentages of the material).

For each of the grading curves, we created two samples (S1 and S2) of aggregate particles. Each sample consists of a total mass of 5\;kg of aggregate material and is carefully designed according to the grain size distribution in Fig.~\ref{fig:histograms} by sieving the raw material in order to separate the different grain size fractions first, and subsequently, by composing the samples according to the dedicated mass-percentages of the size distributions. 

For data acquisition, we made use of a static setup for which the samples are placed in a measurement vessel 
 equipped with a set of calibrated reference markers whose object coordinates are known and which are assembled in a way that they form a common plane with the surface of the aggregate sample. 
We acquired the data by taking images of the aggregate samples (and the reference markers)  which are filled in the the measurement vessel and whose constellation within the vessel is perturbed between the acquisition of each image in order to obtain variations in the sample's visual appearance. 
This acquisition strategy allows to record multiple different images for the individual grading curves by reusing the same sample, consequently reducing the labour-intensive part of material sieving and sample generation.
In this way, we acquired a data set of 900 images in total, consisting of 50 images of each of the two samples (S1 and S2) which were created for each of the nine grading curve definitions, respectively ($50\cdot 2 \cdot 9 = 900$). 

\begin{figure}[ht]
\centering
\includegraphics[width=1.0\columnwidth]{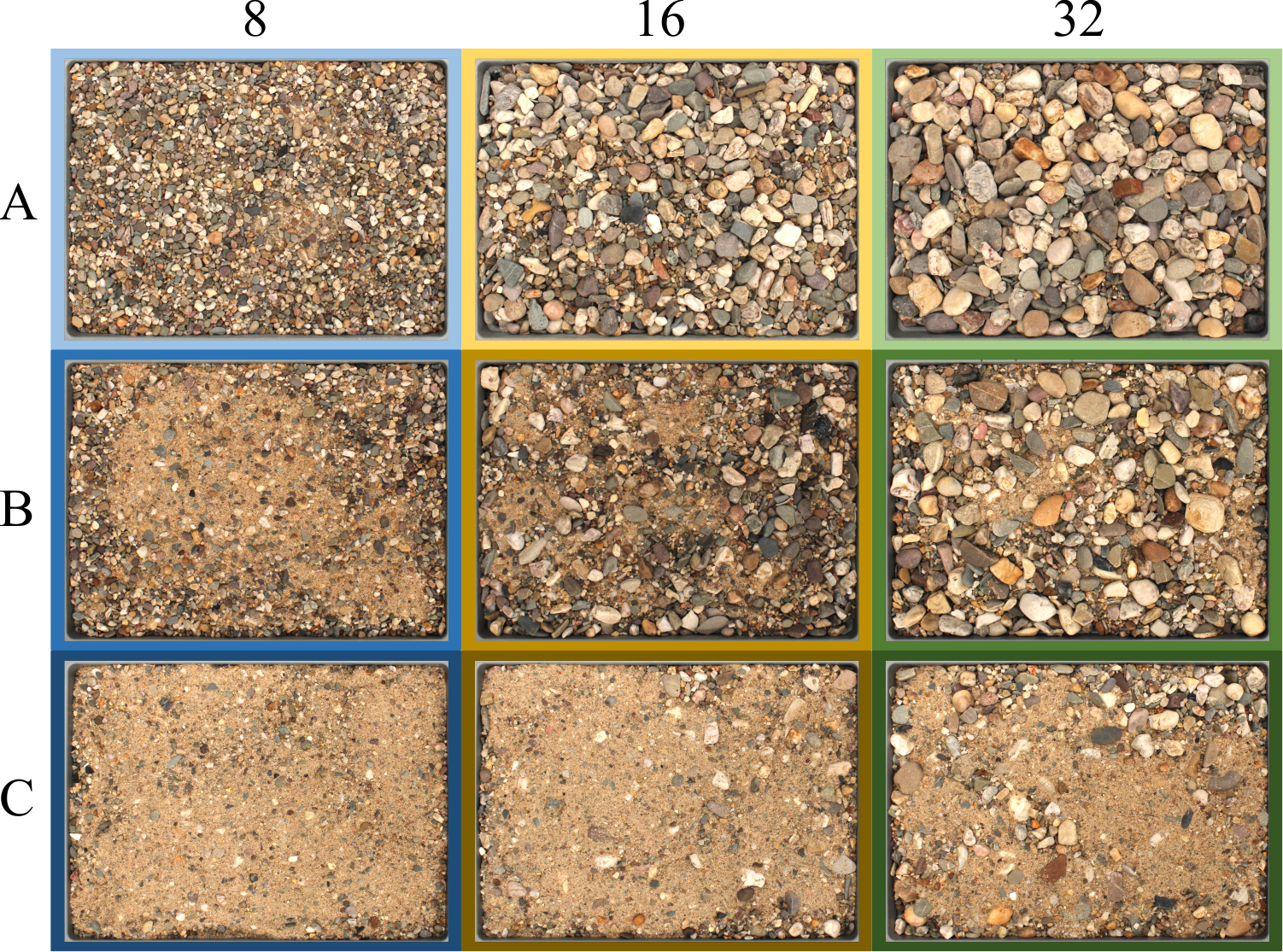}
\caption{Example images of our data set for each of the nine distinguished grading curve classes. The colour-code of the background corresponds to the colour-code of the diagrams in Fig.~\ref{fig:histograms} for an easier visual association.}
\label{fig:examples} 
\end{figure}

For each image, we automatically detect the reference markers, thus receiving the image coordinates of each marker in addition to its known object coordinates. 
We make use of these correspondences for the computation of the homography which describes the perspective transformation of the reference marker's plane in object space (which corresponds to the surface plane of the aggregate sample) to the image plane. 
Using the computed homography, we transform the image in order to obtain an perspectively rectified representation of the aggregate sample with a known, and especially a for the entire image consistent, ground sampling distance (GSD). 
In Fig.~\ref{fig:examples}, example images of our data set showing aggregate samples of each of the distinguished grading curve classes are depicted.
%
%

\subsection{Evaluation strategy and training} \label{sec:evalstrategy}
In order to evaluate the performance of our approach, we make use of the proposed data set and derive suitable performance metrics by comparing the obtained classification results to the given reference information of the grading curves.
We make use of the experimental part of this paper (cf. Sec.~\ref{sec:evaluation}) to investigate the following research questions.
\begin{quote}
\textit{1) How does the image scale used for the input data affect the classification ability of the network?} \textbf{(cf.\;Sec.~\ref{sec:gsd})}
\end{quote}
We analyse the effect of different ground sampling distances (GSD) on the classification results in order to derive guidelines for a minimum required image scale for the differentiation of grading curves. 
\begin{quote}
\textit{2) Which effect do the different constituents of our method, namely the proposed multi-scale encoder (msEnc) and the extensive data augmentation, have on the performance of our approach?} \textbf{(cf.\;Sec.~\ref{sec:ablation})}
\end{quote}
In an ablation study, we investigate the performance differences obtained by training network variants with (\textbf{AggNet:MS}) and without (\textbf{AggNet:Base}) using the proposed msEnc modules, as well as with and without applying the data augmentations described in Sec.~\ref{sec:training}.
In case of the network variant without using the multi-scale feature extraction (AggNet:Base), we replace the convolutional layers in the msEnc-module which have dilation rates larger than 1 (cf. Fig.~\ref{fig:msEnc}) by convolution layers with a dilation rate of 1. 
In this way, both network variants keep the same overall network structure and contain an identical number of weight parameters, allowing a fair comparison between both variants and enabling to evaluate the effect of the multi-scale feature extraction.
\begin{quote}
\textit{3) How well can different and individual grading curves of concrete aggregate be visually distinguished by our proposed method? How can potential errors and limitations be explained and interpreted?} \textbf{(cf.\;Sec.~\ref{sec:interpretation})}
\end{quote}
In this context, we analyse the distinguishability of individual classes, i.e.\;if certain classes (grading curves) can be distinguished more easily compared to others and if so, if an explanation can be found for this behaviour.
Furthermore, to put the achieved results into perspective, we asked human experts to manually classify the test images of our data set and compare the performance of the automatic classification to the results generated by the experts.
Potentially matching patterns in the classification errors made by human experts and the CNN are used to derive an interpretability and explainability of the achieved results and of potential reasons for errors made.

\subsubsection*{Training}
For the evaluation procedure, we comply with the following strategy.
For training and testing our proposed framework, we split the data into a training, validation, and test set. 
More specifically, we make use of all images of the S1 sample set for training of the network, splitting them into 396 actual training images (44 of each class) and 54 validation images (6 of each class).
All images of the S2 sample set are used for testing.
Important to note is, that splitting the images of the two different sample sets into training and test sets is necessary in order to prevent the network to learn to identify individual particles of the sample and their class associations, a potential effect which poses a risk when images of the same sample are used for both, training and testing. 
Each network that is investigated in Sec.~\ref{sec:evaluation} is trained using the described training and validation split with a batch-size of 12 and an initial learning rate of 0.01 which is decreased by a factor of $10^{-1}$ after 10 epochs with no improvements in the training loss. 
We apply early stopping when the training showed no improvement in the validation loss for 10 subsequent epochs and select the network weights with the smallest validation loss as final result.
Furthermore, we apply weight regularisation using L2 penalty with a regularisation factor of $10^{-5}$. 

\subsubsection*{Metrics}
For the quantitative evaluation and for each variant that is to be tested, we trained the respective network five times from scratch and report the average values and standard deviations for the quality metric.
More specifically, we report the average of the overall classification accuracy (OA) and its standard deviation. 
Furthermore, we collect the classification results of the five respective models in a joint confusion matrix and compute the classwise quality according to
\begin{equation}
\text{Quality [\%]} = \frac{TP}{TP+FN+FP}.
\end{equation}
Here, TP (\textit{true positive}) denotes the number of correctly classified images per class, FP (\textit{false positive}) is the number of images that are erroneously classified as the class under consideration, and FN (\textit{false negative}) is the number of images that are erroneously classified as belonging to another class.
In case of the human expert classification, 16 experts were asked to manually associate a grading curve class to the test images. The average OA is computed and the results are gathered in an overall confusion matrix.

\section{Evaluation} \label{sec:evaluation}
In this section, we report the results achieved by the approach proposed in this paper. We analyse and discuss the results w.r.t.\;the research questions of Sec.~\ref{sec:evalstrategy}.

\subsection{Influence of the image resolution} \label{sec:gsd}
\begin{figure}[ht]
\centering
\includegraphics[width=0.99\columnwidth]{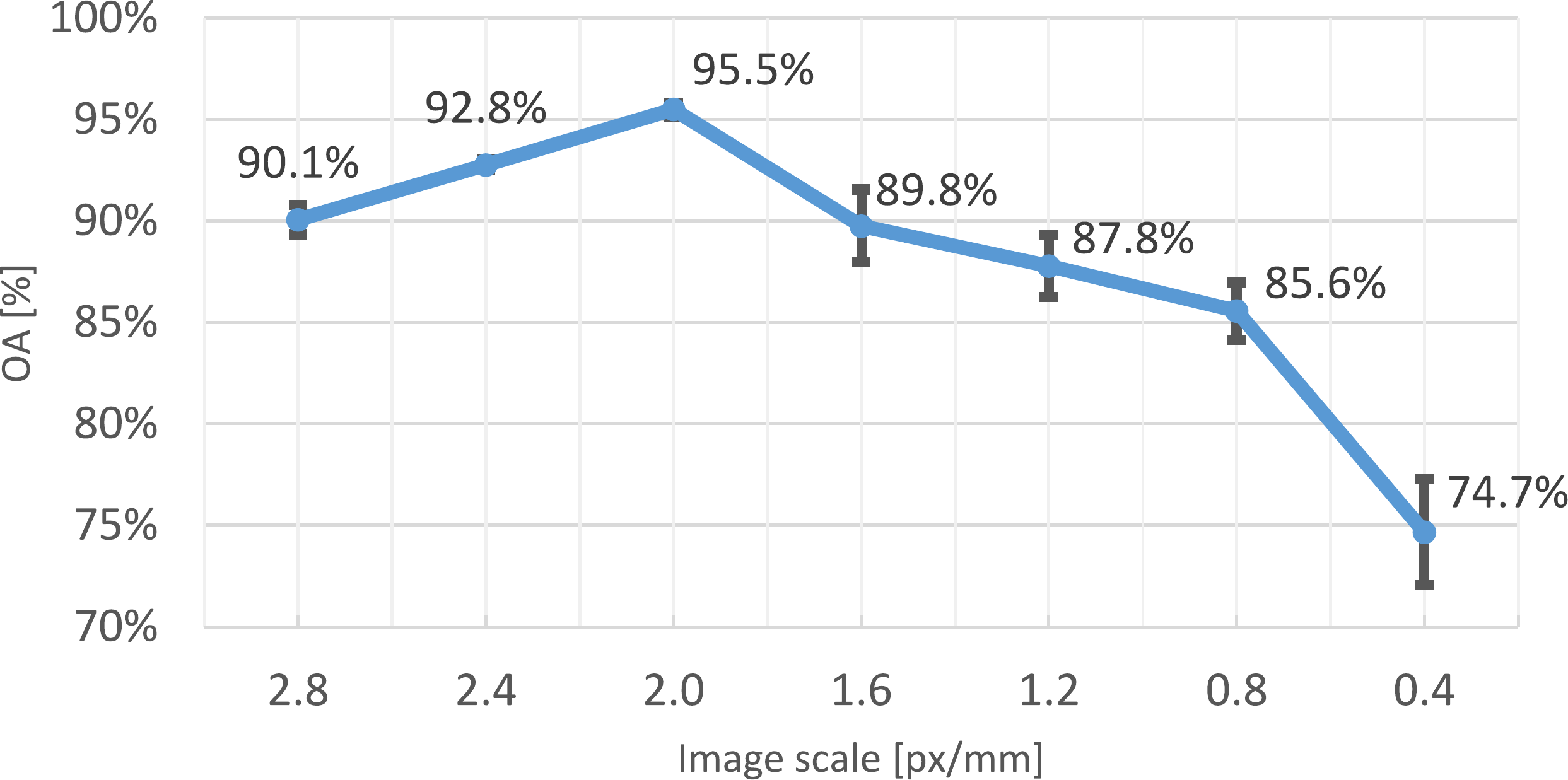}
\caption{Results achieved at different image scales.}
\label{fig:GSD} 
\end{figure}

In order to analyse the classification performance in dependency on the scale of the input images, we trained various models of our network using different ground sampling distances. 
Fig.~\ref{fig:GSD} shows the average OA and its standard deviation obtained for image scales ranging from 0.4\;px/mm to 2.8\;px/mm.

%
As is visible, increasing the image scale from 0.4\;px/mm to 2.0\;px/mm leads to an improvement in the OA of more than 20\%. 
As a conclusion, in this range of image resolutions, increasing the level of detail contained in the images leads to a better performance of the classifier. 
However, as is also visible in Fig.~\ref{fig:GSD}, after an image scale of 2\;px/mm, further increasing the resolution to 2.8\;mm stepwise reduces the mean OA by 5.4\% in total. 
We believe that this counterintuitive behaviour can be caused by the increased spatial size of the images which comes along with increasing the image scale. For instance, while the images of the proposed data set have a spatial size of $750 \times 550$\;px at an image scale of 2\;px/mm, their spatial size is $1050 \times 770$\;px at a scale of 2.8\;px/mm.
As a consequence of that, the effective receptive field of the network and its kernels w.r.t.\;the aggregate area becomes smaller with larger image sizes. 
Together with the fact that in the architecture proposed in this paper, no fully-connected layer, which would incorporate the global image information into the prediction, but instead a global average pooling layer is used at the top of the network, the recepetive field of view of the network is believed to act as the limiting factor w.r.t\;the maximum applicable image size and resolution. 
To overcome this limiting factor, an additional \textit{msEnc}-module can be added to extend the network and its receptive field.
However, we regard the investigation on the influence of adapting the network architecture as out-of-scope of this paper and do not provide an analysis of this effect here.

\subsection{Ablation studies} \label{sec:ablation}
For the remainder of the experiments, an image scale of 2\;px/mm is used, which lead to the best performance of the network according to Fig.~\ref{fig:GSD}. 
The results of the different model variants for the average OA, its standard deviation, as well as the classwise quality scores are reported in Tab.~\ref{tab:Results}.

\begin{table*}[ht]
	\centering
	\caption{Results for average OA and its standard deviation as well as for the classwise quality obtained for different settings of our proposed method (AggNet:Base and AggNet:MS), trained with and without data augmentation, as well as obtained from human expert classification. In all cases, a GSD of 2 [px/mm] was used. The colour coding distinguishes the three largest (green), lowest (red), and intermediate (yellow) values achieved for the quality of the indiviudal classes.}
		\begin{tabular}{l |c|| c c || c c c | c c c | c c c} 
		& Aug. & OA & $\sigma_{OA}$ & \multicolumn{9}{|c}{Classwise quality [\%]} \\ 
		\textit{GSD = 2 [px/mm]} & &[\%] &[\%] & A8 & A16 & A32 & B8 & B16 & B32 & C8 & C16 & C32 \\ \hline
		\vspace{-0.2cm}& & & & & & & & & & & & \\
		AggNet:Base  	&	no	& 73.1 & 4.1 & \textcolor{yellow}{65.4}&  \textcolor{red}{40.5}& \textcolor{green}{81.6}& \textcolor{green}{74.9}& \textcolor{yellow}{47.9}& \textcolor{red}{29.4}& \textcolor{green}{98.0}& \textcolor{yellow}{61.8}& \textcolor{red}{43.0} \\
		AggNet:MS  &	no	& 75.4 & 3.8 & \textcolor{yellow}{69.4}&	\textcolor{red}{51.9}&	\textcolor{green}{76.7}&	\textcolor{green}{80.8}&	\textcolor{yellow}{53.9}&	\textcolor{red}{36.8}&	\textcolor{green}{99.2}&	\textcolor{yellow}{55.8}&	\textcolor{red}{41.7} \\ \hline
		\vspace{-0.2cm}& & & & & & & & & & & & \\
		AggNet:Base  &	yes	& 93.0 & 2.1 & \textcolor{green}{95.6}&	\textcolor{red}{85.6}&	\textcolor{yellow}{90.5}&	\textcolor{green}{92.4}&	\textcolor{yellow}{89.3}&	\textcolor{red}{75.3}&	\textcolor{green}{98.8}&	\textcolor{yellow}{88.5}&	\textcolor{red}{71.6} \\ 
		AggNet:MS & yes	& \textbf{95.5} & \textbf{0.7} & \textcolor{yellow}{94.2}&	\textcolor{red}{88.1}&	\textcolor{green}{92.9}&	\textcolor{green}{95.2}&	\textcolor{yellow}{94.7}&	\textcolor{red}{83.8}&	\textcolor{green}{98.8}&	\textcolor{yellow}{93.4}&	\textcolor{red}{83.0} \\ \hline
		\vspace{-0.2cm}& & & & & & & & & & & & \\
		Expert classification & - & 62.8 &11.0 & \textcolor{yellow}{42.9}&	\textcolor{red}{26.9}&	\textcolor{green}{66.9}&	\textcolor{yellow}{47.9}&	\textcolor{yellow}{40.8}&	\textcolor{red}{32.9}&	\textcolor{green}{70.1}&	\textcolor{red}{28.9}&	\textcolor{green}{57.9}\\ \hline		
		\end{tabular}	
\label{tab:Results}
\end{table*}  
The average OA that is achieved by our proposed method (AggNet:MS with augmentation) is 95.5\% $\pm$ 0.7\%.
Compared to the OA achieved by the manual classification of the human experts (62.8\%), the automatic classification by our approach leads to significantly better results. 
What is also visible from Tab.~\ref{tab:Results} is the large effect, that the applied data augmentation has on the classification performance. 
For both network variants (AggNet:Base and AggNet:MS), applying data augmentation leads to an improvement of the average OA of about 20\%. 

Furthermore, the proposed multi-scale feature extraction leads to an improvement in the average OA of more than 2\% for both, training with and without data augmentation.
However, this improvement has to be seen in relation to the relatively large standard deviations of the OA, which ranges from 0.7\% to 4.1\%. 
Still, the consistent improvements of the OA indicates the desired beneficial effect of the proposed msEnc-modules for the classification of grading curves with particles of strongly varying sizes. 
Having a closer look on the classwise quality metrics achieved by the network with and without the multi-scale feature extraction (AggNet:Base and AggNet:MS) with applied data augmentation (right part of Tab.~\ref{tab:Results}), it becomes visible, that the largest improvements of the AggNet:MS model compared to the AggNet:Base model are achieved for classes containing particles of the fractions 16\;mm and 32\;mm. 
While the improvements of the classwise quality of the grading curve classes containing a largest grain of 8\;mm is at maximum 2.8\% (for the class B8), the improvements for the classes with a largest grain of 16\;mm and 32\;mm are significantly higher and range from 2.4\% (class A32) up to 11.4\% (class C32).
These observations support our argumentation in Sec.~\ref{sec:architecture} w.r.t.\;the multi-scale design of the proposed msEnc-modules, being beneficial especially in cases where the images contain a larger range of object sizes (0-16\;mm and 0-32\;mm, respectively). 

\subsection{Interpretation} \label{sec:interpretation}

The colour coding of the classwise quality metrics in Tab.~\ref{tab:Results}, distinguishing the three highest/middle/lowest quality values (green/yellow/red) achieved by the different models, reveals a pattern that is (almost) identical for all model variants and also for the manual human classification. 
It can be seen, that the two extrema of the grain size distributions (class C8: very fine containing only the smallest particle fractions and class A32: very coarse containing large amounts of particles of the maximum grain size fraction) are always among the classes with the best quality achieved by the automatic as well as the manual classification. An exception is the AggNet:Base variant without data augmentation, where the class A32 achieved the fourth best quality value, but is with 90.5\% still high.
On the contrary side, the classes with the lowest quality achieved (A16, B32, C32), are the classes in the intermediate area of the grain size distributions, which contain a comparable amount of particles from the whole (B32 and C32) or at least a broad (A16) range of the particle fractions (see also Fig.~\ref{fig:hist01}, where the relatively uniform composition of these classes with the different fractions is visible).
The observations above allow the conclusion, that the extrema of the grain size distribution are visually well distinguishable by both, the CNN and humans, while grading curves lying in the middle range of the size distributions are the ones which are most difficult to distinguish.

\begin{figure}[ht]
\centering
\subfloat[Expert classification.] {\includegraphics[width=0.95\columnwidth]{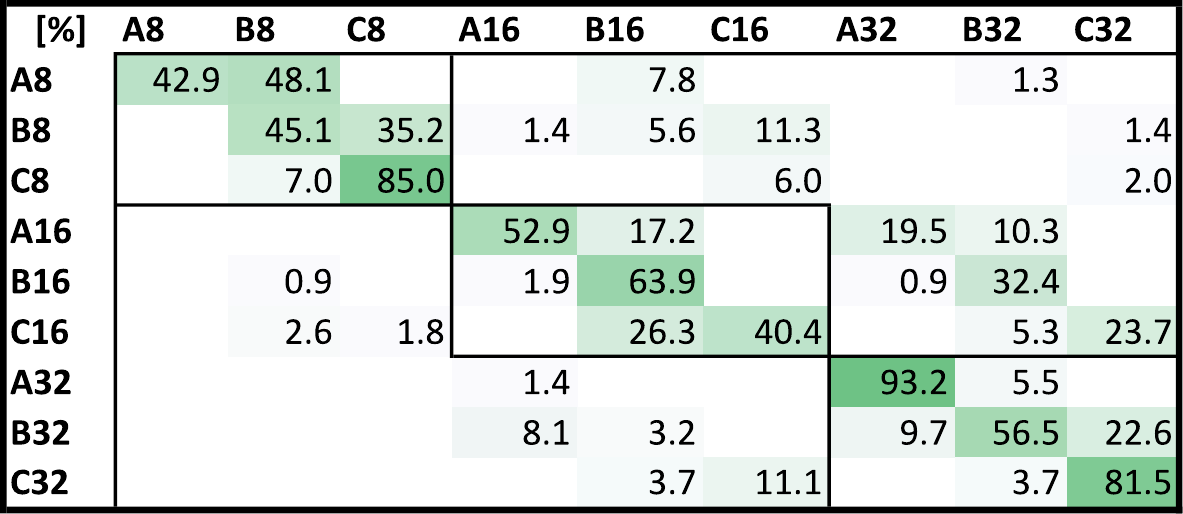}\label{fig:ConfMatExpert}} 

\subfloat[AggNet:MS (without augmentation).] {\includegraphics[width=0.95\columnwidth]{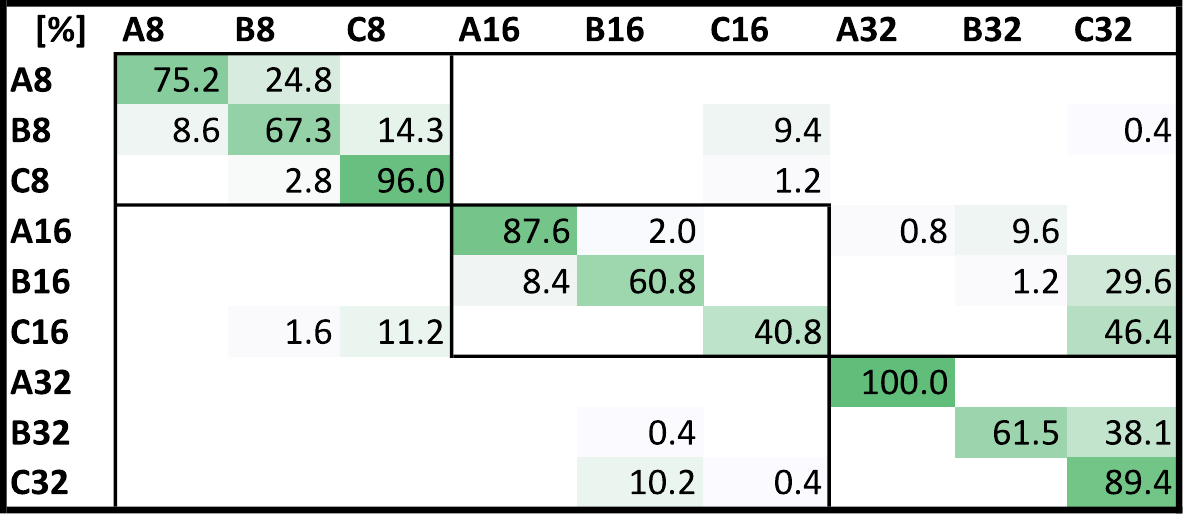}\label{fig:ConfMatAggMSNoAug}} 

\subfloat[AggNet:MS (with augmentation).] {\includegraphics[width=0.95\columnwidth]{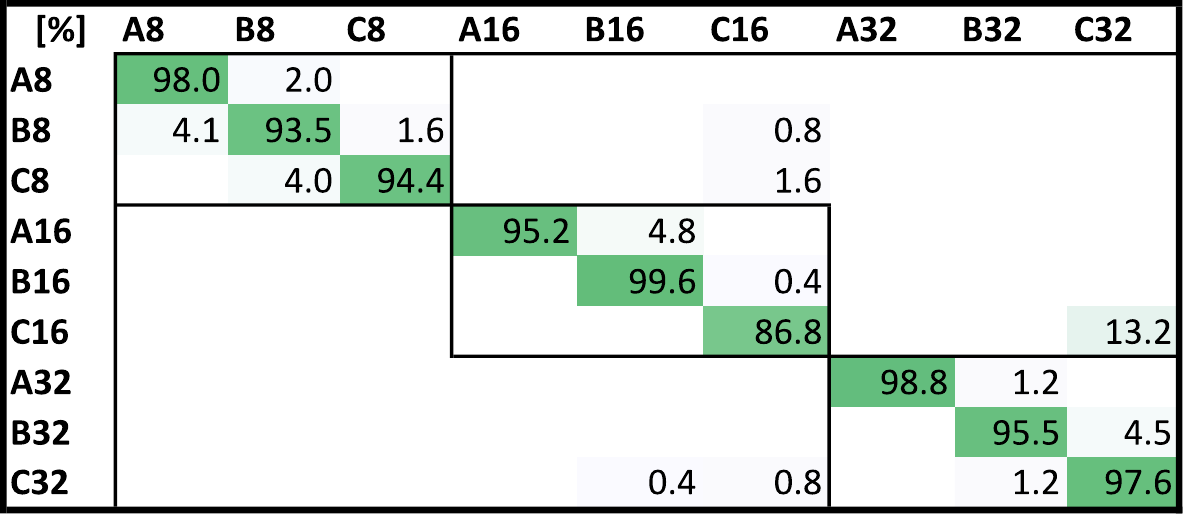}\label{fig:ConfMatAggMSAug}} 
\caption{Confusion matrices obtained from different classifications. The values are to be read as amount of images (in [\%]) of each reference class (rows) that are classified as the predicted class (columns). The colour coding denotes low values (light green) and high values (dark green). Empty entries correspond to values 0.0.}
\label{fig:ConfMats} 
\end{figure}

A closer look on the error distributions resulting from the expert classification as well as from the AggNet:MS with and without data augmentation is given by the confusion matrices in Fig.~\ref{fig:ConfMats}.
%
%
Here, the numbers are to be read as how many images (in [\%]) of a certain reference class (rows) are classified as the predicted class (columns). 
The confusion matrix of the expert classification (Fig.~\ref{fig:ConfMatExpert}) reveals a clear pattern. 
Confusions are mainly located either in directly neighbouring classes, i.e.\;grading curves containing particles of the same size fractions (8, 16, 32) but with a different granularity (A, B, C), visible from the entries directly next to the main diagonal, or are shifted towards classes of the same granularity (A, B, C) but with a different largest grain (8, 16, 32), visible from the minor diagonals that are shifted parallel to the main diagonal. 
In other words: Errors made by the human annotators almost exclusively occur from confusing either the granularity (fine vs.\;medium fine and medium fine vs.\;coarse) or from confusing the largest grain (8\;mm vs.\;16\;mm and 16\;mm vs.\;32\;mm).
The same pattern is also visible in the confusion matrices obtained from the classifications made by the proposed CNN (cf.\;Fig.~\ref{fig:ConfMatAggMSNoAug} and \ref{fig:ConfMatAggMSAug}), although less distinct since the overall amount of erroneous classifications is significantly less. 
This behaviour leads to the conclusion, that the CNN based classification suffers from the same effect as the human differentiation of the grading curves, namely that on the one hand, grading curves containing the same particle size fractions but with a different proportion, may look very much alike and, on the other hand, that the identification of the largest size fraction is difficult in some cases. 
While for the AggNet:MS model trained without augmentation, both of the factors just described are valid (cf.\;Fig.~\ref{fig:ConfMatAggMSNoAug}), the usage of data augmentation during training almost completely eliminates the confusions between grading curves being composed of different size fractions, as most remaining errors appear as confusions between granularities (cf.~\ref{fig:ConfMatAggMSAug}).
Having said that, a larger amount of confusions, namely 13.2\%, still remains between the classes C16 and C32, i.e.\;grading curves that contain a huge proportion of both, very fine (0-2\;mm) and fine (2-8\;mm) material, and only small proportions of coarser particles (8-16\;mm and 16-32\;mm, respectively). See the example images of both classes in Fig.~\ref{fig:examples} for a visual impression. 
In many cases, apparently, these small proportions of the coarse fractions are not sufficient enough, to identify the largest fraction contained, therefore leading to the larger amount of confusions between those classes.

\section{Conclusion} \label{sec:conclusion}
In current practice of concrete production, the grading curve of the concrete aggregates is usually determined from small random batch-samples using mechanical sieving, thus extrapolating from a sample of a few kilograms to many tons of aggregates. As a consequence, the unknown variations of the aggregate's grading curve (especially pronounced in the case of recycled materials), is not properly taken into account in the mix design, leading to undesired effects w.r.t.\,the concrete's properties. 
In order to enable the control of the variations, we present a CNN based method for the determination of the grading curve of concrete aggregate from images.
In this context, we present a suitable network architecture proposing a multi-scale encoder module which improves the prediction performance by enabling the feature extraction for scenes with strongly varying object sizes. 
We created and published a data set of concrete aggregate together including high-resolution images and the corresponding reference grading curves.

In the future, we aim at extending the approach by not only determining the discrete grading curve belonging to the aggregate sample but instead, to predict the percentiles of the size distribution directly, allowing a more flexible application of the approach into the process of concrete production.
Furthermore, we aim at applying the proposed approach as a basis for the development of an online concrete control scheme, with the goal to adapt the mix composition in real time to react to the detected fluctuations in the raw materials. 

\section*{ACKNOWLEDGEMENTS}\label{ACKNOWLEDGEMENTS}
This work was supported by the Federal Ministry of Education and Research of Germany (BMBF) as part of the research project ReCyCONtrol [Project number 033R260A]. \\
(\url{https://www.recycontrol.uni-hannover.de}).

\renewcommand{\bibsection}{\section*{REFERENCES}}
{
	\begin{spacing}{1.17}
		\normalsize
		\bibliography{Literatur} 
	\end{spacing}
}

\end{document}